\documentclass{article}

\usepackage[numbers]{natbib}
\usepackage{authblk}
\usepackage{fullpage}

\usepackage[utf8]{inputenc} 
\usepackage[T1]{fontenc}    
\usepackage{hyperref}       
\usepackage{url}            
\usepackage{booktabs}       
\usepackage{nicefrac}       
\usepackage{xcolor}   

\usepackage{authblk}
\usepackage{wrapfig}
\usepackage{microtype}
\usepackage{graphicx}
\usepackage{subfigure}
\usepackage{algorithm}
\usepackage{tabularx}
\usepackage[noend]{algorithmic}

\usepackage{amsfonts}       
\usepackage{amsmath}
\usepackage{amsthm}
\usepackage{dsfont}
\usepackage{amssymb}
\usepackage{amscd}
\usepackage{mathtools}
\mathtoolsset{showonlyrefs=true}
\usepackage{enumerate}
\usepackage{authblk}
\usepackage{ bbold }
\usepackage{thmtools} 
\usepackage{thm-restate}

\title{Offline Reinforcement Learning as Anti-Exploration}

\newcolumntype{a}{>{\hsize=0.75\hsize}X}
\newcolumntype{b}{>{\hsize=1.2\hsize}X}
\newcolumntype{x}{>{\hsize=.42\hsize}X}
\newcolumntype{y}{>{\hsize=.25\hsize}X}

\author[1]{Shideh Rezaeifar\footnote{Equal contribution.}}
\newcommand\CoAuthorMark{\footnotemark[\arabic{footnote}]}
\author[2]{Robert Dadashi\protect\CoAuthorMark}
\author[2,3]{Nino Vieillard}
\author[2,4]{Léonard Hussenot}
\author[2]{Olivier Bachem}
\author[2]{Olivier Pietquin}
\author[2]{Matthieu Geist}
\affil[1]{University of Geneva}
\affil[2]{Google Research, Brain Team}
\affil[3]{Univ. Lorraine, CNRS, Inria, IECL, F-54000 Nancy,
France}
\affil[4]{Univ. Lille, CNRS, Inria Scool, UMR 9189 CRIStAL}

\date{}

\newcommand{\argmax}{\operatorname{argmax}} 

\newcommand{\sm}{\operatorname{softmax}}
\newcommand{\kl}[2]{\operatorname{KL}(#1||#2)}

\newcommand{\R}{\mathbb{R}}
\newcommand{\hc}{\mathcal{H}}

\newcommand{\dataset}{\mathcal{D}}
\newcommand{\bellman}{\mathcal{B}}
\newcommand{\states}{\mathcal{S}}
\newcommand{\actions}{\mathcal{A}}

\newcommand{\E}{\mathbb{E}}
\newcommand{\U}{\mathbb{U}}

\newcommand{\norm}[1]{\left\lVert#1\right\rVert}

\begin{document}

\maketitle

\begin{abstract}
    \looseness=-1
  Offline Reinforcement Learning (RL) aims at learning an optimal control from a fixed dataset, without interactions with the system. An agent in this setting should avoid selecting actions whose consequences cannot be predicted from the data. This is the converse of exploration in RL, which favors such actions. We thus take inspiration from the literature on bonus-based exploration to design a new offline RL agent. The core idea is to subtract a prediction-based exploration bonus from the reward, instead of adding it for exploration. This allows the policy to stay close to the support of the dataset. We connect this approach to a more common regularization of the learned policy towards the data. Instantiated with a bonus based on the prediction error of a variational autoencoder, we show that our agent is competitive with the state of the art on a set of continuous control locomotion and manipulation tasks.
\end{abstract}

\section{Introduction}
\looseness=-1
Deep Reinforcement Learning (RL) has achieved remarkable success in a variety of tasks including robotics \citep{kober2013reinforcement, thrun1995approach, benbrahim1997biped, bagnell2001autonomous, endo2008learning}, recommendation systems \citep{rojanavasu2005new, afsar2021reinforcement}, and  games \citep{silver2018general}. Deep RL algorithms generally assume that an agent repeatedly interacts with an environment and uses the gathered data to improve its policy: they are said to be \emph{online}. Because actual interactions with the environment are necessary to online RL algorithms, they do not comply with most of real-world applications constraints. Indeed, allowing an agent to collect new data may be infeasible such as in healthcare~\citep{murphy2001marginal}, autonomous driving \citep{sallab2017deep, grigorescu2020survey}, or education \citep{mandel2014offline}. As an alternative, \emph{offline} RL~\cite{levine2020offline} is a practical paradigm where an agent is trained using a fixed dataset of trajectories, without any interactions with the environment during learning. The ability to learn from a fixed dataset is the crucial step towards scalable and generalizable data-driven learning methods.

In principle, off-policy algorithms \citep{ernst2005tree,lagoudakis2003least,mnih2013playing, lillicrap2019continuous, haarnoja18b} could be used to learn from a fixed dataset. However, in practice, they perform poorly without any feedback from the environment. This issue persists even when the off-policy data comes from effective expert policies, which in principle should address any exploration issue \cite{kumar2019stabilizing}. The main challenge comes from the sensitivity to the data distribution. The distribution mismatch between the behavior policy and the learned policy leads to an extrapolation error of the value function, which can become overly optimistic in areas of state-action space outside the support of the dataset.  The extrapolation error  accumulates along the episode and results in unstable learning and divergence \citep{fujimoto2018addressing, kumar2019stabilizing, cql, levine2020offline, peng2019advantage, dadashi2021offline, yu2020mopo}.

This work introduces a new approach to Offline RL, inspired by exploration. This may seem counter-intuitive. Indeed, in online RL, an exploring agent will try to visit state-action pairs it has never experienced before, hence drifting from the distribution in the dataset. This is exactly what an Offline RL agent should avoid, so we frame it as an \emph{anti-exploration} problem. 
We focus on bonus-based exploration~\citep{brafman2002r, burda2018exploration, pathak2017curiosity, burda2018large, BellemareSOSSM16}. The underlying principle, in these methods, is to add a bonus to the reward function, this bonus being higher for novel/surprising state action-pairs \citep{barto2013novelty}. The core idea of the proposed approach is to perform \emph{anti-exploration} by \emph{subtracting} the bonus from the reward, instead of adding it, effectively preventing the agent from selecting actions of which the effects are unknown.
Under minimal assumptions, we relate this approach to the more common offline RL idea  of penalizing the learned policy for deviating from the policy that supposedly generated the data. We also propose a specific instantiation of this general idea, using TD3~\citep{fujimoto2018addressing} as the learning agent, and for which the bonus is the reconstruction error of a variational autoencoder (VAE) trained on offline state-action pairs. We evaluate the agent on the hand manipulation and locomotion tasks of the D4RL benchmark \citep{d4rl}, and show that it is competitive with the state of the art.

\section{Preliminaries}
\label{sec:background}
\looseness=-1
A Markov decision process (MDP) is defined by a tuple $\mathcal{M} := (\states, \actions, r, P, \gamma)$, with $\states$ the state space, $\actions$ the action space,  $P \in\Delta_\states^{\states\times\actions}$ the Markovian transition kernel ($\Delta_\states$ is the set of distributions over $\states$), $r\in\mathbb{R}^{\states\times\actions}$ the reward function and $\gamma\in(0,1)$ the discount factor.  A policy $\pi\in\Delta_\actions
^\states$ is a mapping from states to distribution over actions (a deterministic policy being a special case). The general objective of RL is to find a policy that maximizes the expectation of the return $G_t = \sum^{\infty}_{i=0}  \gamma^i r(s_{t +i}, a_{t + i})$. 
Given a policy $\pi$, the action value function, or Q-function, is defined as the expected return when following the policy $\pi$ after taking action $a$ in state $s$, $Q^\pi(s, a) =  \E_{\pi}[G_t| s_t = s, a_t = a]$.

For $u_1,u_2 \in \R^{\states\times\actions}$, define the dot product $\langle u_1, u_2 \rangle =\sum_{a} u_1(\cdot,a) u_2(\cdot,a)\in\R^\states$, and for $v\in\R^\states$ write $Pv = \sum_{s'} P(s'|\cdot,\cdot) v(s')\in\R^{\states\times\actions}$. Whith these notations, define the Bellman operator as $\bellman^\pi Q = r + \gamma P \langle \pi, Q\rangle$. It is a contraction~\citep{puterman2014markov} and $Q^\pi$ is its unique fixed point. An optimal policy $\pi^*$ satisfies $\pi^*\in\argmax_\pi Q^\pi$. A policy $\pi$ is said to be a greedy policy with respect to $Q$ if $\pi\in\argmax_{\pi'} \langle \pi, Q\rangle$ (notice that $\langle \pi, Q\rangle(s)=E_{a\sim\pi(\cdot|s)}[Q(s,a)]$). An algorithm that allows computing an optimal policy is Value Iteration (VI). For any initial $Q_0\in\R^{\states\times\actions}$, it iterates as follows,
\begin{equation}
    \begin{cases}
        \pi_{k+1} \in\argmax_{\pi\in\Delta_\actions^\states} \langle \pi, Q_k\rangle 
        \\
        Q_{k+1} = r + \gamma P\langle \pi_{k+1}, Q_k\rangle
    \end{cases}.
\end{equation}
The first step is the greedy step. When considering only deterministic policies (slightly abusing notations), it simplifies to $\pi_{k+1}(s) = \argmax_{a\in\actions} Q_k(s,a)$. The second step is the evaluation step, again for deterministic policies, it simplifies to $Q_{k+1}(s,a) = r(s,a) + \gamma \E_{s'|s,a}[\max_{a'}Q_k(s',a')]$.

VI may be used to derive many existing deep RL algorithms. First, consider the discrete action case, assume that a dataset of collected transitions $\mathcal{D} = \{(s,a,s',r)\}$ is available. The seminal DQN~\citep{mnih2013playing} parameterizes the $Q$-value with a deep neural network $Q_\omega$, takes $Q_k$ as being a copy of a previous network $Q_{\bar{\omega}}$ ($\bar{\omega}$ being a copy of the parameters), uses the fact that the greedy policy can be exactly computed, and considers a loss which is the squared difference of both sides of the evaluation step, $L_\text{dqn}(\omega) = \hat{\E}_\mathcal{D}[(r + \gamma \max_{a'} Q_{\bar{\omega}}(s',a') - Q_\omega(s,a))^2]$ , with $\hat{\E}$ an empirical expectation.

With continuous actions, the greedy policy can no longer be computed exactly, which requires introducing a policy network $\pi_\theta$. TD3~\citep{fujimoto2018addressing}, a state-of-the-art actor-critic algorithm, can be derived from the same VI viewpoint. It adopts a fixed-variance Gaussian parameterization for the policy, $\pi_\theta\sim\mathcal{N}(\mu_\theta, \sigma I)$ with $\mu_\theta\in\actions^\states$, $\sigma$ the standard deviation of the exploration noise and $I$ the identity matrix. The greedy step can be approximated by the actor loss $J_{\text{td3, actor}}(\theta) = \hat{\E}_{s\in\mathcal{D}}[\hat{\E}_{a\sim\pi_\theta(\cdot, s)}[Q_{\bar{\omega}}(s,a)]]$. This can be made more convenient by using the reparameterization trick, $J_{\text{td3, actor}}(\theta) = \hat{\E}_{s\in\mathcal{D}}[\hat{\E}_{\epsilon\sim\mathcal{N}(0,\sigma I)}[Q_{\bar{\omega}}(s,\mu_\theta(s) + \epsilon)]]$. The critic loss is similar to the DQN one, $J_{\text{td3, critic}}(\omega) = \hat{\E}_{\mathcal{D}}[\hat{E}_{a'\sim\pi_\theta(\cdot|s')}[(r + \gamma Q_{\bar{\omega}}(s',a') - Q_\omega(s,a))^2]$.

\section{Framing offline RL as anti-exploration}
\label{sec:bonus}

\textbf{What makes offline reinforcement learning hard?} Conventional on-policy RL algorithms operate in a setting where an agent repeatedly interacts with the environment, gathers new data and uses the data to update its policy.
The term off-policy denotes an algorithm that can use the data collected by other policies whilst still being able to interact with the environment. Offline RL relies solely on a previously collected dataset without further interaction with the environment. This setting can leverage the vast amount of previously collected datasets, \emph{e.g.} human demonstrations or hand-engineered exploration strategies. 
Offline RL is challenging because the collected dataset does not cover the entire state-action space. The out-of-distribution (OOD) actions can cause an extrapolation error in the value function approximations. As an example, consider the regression target of the DQN loss, $y=r+\gamma \max_{a'\in\actions} Q_{\bar{\omega}}(s',a')$. The estimate of the value function $Q_{\bar{\omega}}(s',a')$ for state-action pairs that are not in the dataset can be erroneously high due to extrapolation errors. As a consequence, the maximum  $\max_{a'\in\actions} Q_{\bar{\omega}}(s',a')$ may be reached for a state-action couple $(s',a')$ that has never been observed in the data. Using this maximal value as part of the target for estimating $Q_\omega(s,a)$ will result in being over-optimistic about $(s,a)$. This estimation error accumulates with time as the policy selects actions that maximize the value function. Thus, many methods constrain the learned policy to stay within the support of the dataset. These methods differ in how the deviation is measured and how the constraint is enforced. For example, it could be achieved by modifying the DQN target to $r + \gamma \max_{a'|(s',a')\in\mathcal{D}} Q_\omega(s',a')$, which is the underlying idea  of~\citep{kumar2019stabilizing,ghasemipour2020emaq}, among others.

\textbf{Exploration in Reinforcement Learning.} 
Exploration is critical in RL and failing at handling it properly can prevent agents to identify high reward regions, or even to gather any reward at all if it is sparse. There are many approaches to exploration, as well as many challenges, such as the well-known  exploration-exploitation dilemma. In this paper, we focus on bonus-based exploration and then adapt it to offline RL. The core idea is to define or learn a bonus function $b\in\R^{\states\times\actions}$, which is small for known state-action pairs and high for unknown ones. This bonus is \emph{added} to the reward function which will intuitively drive the learned policy to follow trajectories of unknown state-action pairs. Embedded within a generic VI approach, this can be written as
\begin{equation}
    \begin{cases}
        \pi_{k+1} \in\argmax_{\pi\in\Delta_\actions^\states} \langle \pi, Q_k\rangle 
        \\
        Q_{k+1} = r {\color{blue}+ b} + \gamma P\langle \pi_{k+1}, Q_k\rangle
    \end{cases}.
\end{equation}

\looseness=-1
The bonus-based strategies can be roughly categorized into count-based or prediction-based methods. 
First, in count-based methods, the novelty of a state is measured by the number of visits, and a bonus is assigned accordingly. For example, it can be $b(s,a)\propto 1/\sqrt{n(s,a)}$, with $n(s,a)$ the number of times the state-action couple has been encountered. The bonus guides the agent’s behavior to prefer rarely visited states over common ones. When the state-action space is too large, counting is not a viable option, but the frequency of visits can be approximated by using a density model~\citep{BellemareSOSSM16, ostrovski2017count} or mapping state-actions to hash codes~\citep{tang2017exploration}.
Second, in prediction-based methods, the novelty is related to  the agent’s knowledge about the environment. The agent’s familiarity with the environment can be estimated through a prediction model \citep{achiam2017surprise, Schmidhuber1991, pathak2017curiosity, burda2018large}.  For example, a forward prediction model captures an agent's ability to predict the outcome of its own actions. High prediction error indicates that the agent is less familiar with that state-action pair and vice versa. However, this prediction error is entangled with the agent's performance in the specific task. Random network distillation (RND) was introduced as an alternative method where the prediction task is random and independent of the main one \citep{burda2018exploration}.

\textbf{Proposed anti-exploration approach.} 
Offline RL algorithms learn from a fixed dataset without any interaction with the environment. State-action pairs outside of the dataset are therefore never actually experienced and can receive erroneously optimistic value estimates because of extrapolation errors that are not corrected by an environment feedback. The overestimation of value functions can be encouraged in online reinforcement learning, to a certain extent, as it incentivizes agents to explore and learn by trial and error \citep{gulcehre2021regularized, Schmidhuber1991}. Moreover, in an online setting, if the agent wrongly assigns a high value to a given action, this action will be chosen, the real return will be experimented, and the value of the action will be corrected through bootstrapping. In this sense, online RL is self-correcting. On the contrary, in offline RL, the converse of exploration is required to keep the state-actions close to the support of the dataset (no self-correction is possible given the absence of interaction).  Hence, a natural idea consists in defining an \emph{anti-exploration} bonus to penalize OOD state-action pairs. This bonus will encourage the policy to act similarly to existing transitions of the offline trajectories. 

A naive approach to anti-exploration for offline RL would thus consist in \emph{subtracting} the bonus from the reward, instead of adding it. Embedded within our general VI scheme, it would give
\begin{equation}
    \begin{cases}
        \pi_{k+1} = \argmax_\pi \langle \pi, Q_k  \rangle
        \\
        Q_{k+1} = r { \color{red} -  b} + \gamma P \langle \pi_{k+1}, Q_k \rangle
    \end{cases}.
\end{equation}
Intuitively, this should prevent the RL agent from choosing actions with high bonus, i.e., unknown actions that are not or not enough present in the dataset. However, this would not be effective at all. Indeed, in offline RL, we would only use state-action pairs in the dataset, for which the bonus is supposedly low. Hence this would not avoid bootstrapping values of actions with unknown consequences, which is our primary goal. As an example, consider again the bootstrapped target of DQN, with the additional bonus, $r {\color{red} - b(s,a)} + \gamma \max_{a'\in\actions} Q_{\bar{\omega}}(s',a')$. The state action-couple $(s,a)$ necessarily comes from the dataset, so the bonus is supposedly low, for example zero for a well trained prediction-based bonus. Thus, the bonus is here essentially useless. What would make more sense would be to penalize the bootstrapped value with the bonus, for example considering the target $r + \gamma \max_{a'\in\actions} (Q_{\bar{\omega}}(s',a') {\color{red} - b(s',a')})$. This way, we penalize bootstrapping values for unknown state-action pairs $(s',a')$. It happens conveniently that both approaches are equivalent, from a dynamic programming viewpoint (that is, when Q-values and policies are computed exactly for all possible state-action pairs; the equivalence obviously does not hold in an approximate  offline setting). Indeed, considering again our VI scheme, we have
\begin{equation}
        \begin{cases}
        \pi_{k+1} = \argmax_\pi \langle \pi, Q_k  \rangle
        \\
        Q_{k+1} = r  {\color{red} - b} + \gamma P \langle \pi_{k+1}, Q_k \rangle
    \end{cases}
    \Leftrightarrow
    \begin{cases}
        \pi_{k+1} = \argmax_\pi \langle \pi, Q'_k  {\color{red}- b} \rangle
        \\
        Q'_{k+1} = r + \gamma P \langle \pi_{k+1}, Q'_k  {\color{red}  -b} \rangle
    \end{cases},
    \label{eq:ae_vi}
\end{equation}

with $Q'_{k+1} = Q_{k+1} + b$. Even though the $Q$-values are not the same, both algorithms provide the same sequence of policies, provided that $Q'_0 = Q_0 + b$. In fact, independently from the initial $Q$-values, both algorithms will converge to the policy maximizing  $\E_\pi[\sum_t {\gamma^t(r(s_t,a_t) {\color{red}- b(s_t,a_t)})}]$. This comes from the fact that is is a specific instance of a regularized MDP~\cite{geist2019theory}, with a linear regularizer $\Omega(\pi) = \langle \pi, b\rangle$. This is a better basis for an offline agent, as the bonus directly affects the $Q$-function. This was illustrated above with the example of the DQN target as a specific instance of this VI scheme. However, the idea holds more generally, for example within an actor-critic scheme, as we will exemplify later by providing an agent based on TD3 for continuous actions.

\section{A link to regularization}
\label{sec:regualrization}

\looseness=-1
Many recent papers in offline RL focus on regularizing the learned policy to stay close to the training dataset of offline trajectories.It usually amounts to a penalization based on a divergence between the learned policy and the behavior policy (the action distribution conditioned on states underlying the dataset).These methods share the same principle, but they differ notably in how the deviation of policies is measured. Here, we draw a link between anti-exploration and behavior-regularized offline RL.

Let $b\in\R^{\states\times\actions}$ be any exploration bonus. We will just assume that the bonus is lower for state-action pairs in the dataset than for those outside. For example, if $b$ was trained with a one-class classifier, it could be $b(s,a)\approx 0$ if $(s,a)$ in the support, $b(s,a)\approx 1$ elsewhere. We will discuss a different approach later, based on the reconstruction error of a conditional variational autoencoder.

We can use this bonus to model a distribution of actions conditioned on states, assigning high probabilities for low bonuses, and low probabilities for high bonuses, the goal being to model the data distribution (which is a hard problem in general). This can be conveniently done with a softmax distribution. Let $\beta>0$ be a scale parameter and $\tau>0$ a temperature, we define the policy $\pi_b$ as
\begin{equation}
    \pi_b(.|s) = \sm\left(-\frac{\beta}{\tau} b(s,.)\right).
\end{equation}
Now, we can use this policy to regularize a VI scheme. Define the Kullback-Leibler (KL) divergence between two policies as $\kl{\pi_1}{\pi_2} = \langle \pi_1, \ln \pi_1 - \ln \pi_2 \rangle \in\R^\states$. Define also the entropy of a policy as $\hc(\pi) = - \langle \pi, \ln\pi\rangle\in\R^\states$. Consider the following KL-regularized VI scheme~\citep{geist2019theory,vieillard2020leverage}:
\begin{equation}
    \begin{cases}
        \pi_{k+1} = \argmax_\pi \left(\langle \pi, Q_k \rangle - \tau \kl{\pi}{\pi_b}\right)
        \\
        Q_{k+1} = r + \gamma P\left(\langle \pi_{k+1}, Q_k\rangle - \tau\kl{\pi_{k+1}}{\pi_b}\right)
    \end{cases}. \label{eq:regVI_KL}
\end{equation}
Consider for example the quantity optimized within the greedy step, we have that
\begin{equation}
    \langle \pi, Q_k\rangle - \tau \kl{\pi}{\pi_b} = \langle \pi, Q_k + \tau \ln \sm(-\frac{\beta b}{\tau})\rangle - \tau \langle \pi, \ln\pi\rangle = \langle \pi, Q_k - \Delta_b\rangle + \hc(\pi),
\end{equation}
with $\Delta_b(s,a) = \beta b(s,a) + \tau \ln \sum_{a'}\exp(-\beta b(s,a')/\tau$. The same derivation can be done for the evaluation step, showing that 
\begin{equation}
    \text{\eqref{eq:regVI_KL}} \Leftrightarrow
    \begin{cases}
        \pi_{k+1} = \argmax_\pi \left(\langle \pi, Q_k - \Delta_b \rangle + \tau \hc(\pi)\right)
        \\
        Q_{k+1} = r + \gamma P\left(\langle \pi_{k+1}, Q_k - \Delta_b \rangle + \tau\hc(\pi_{k+1})\right)
    \end{cases}.
\end{equation}
This can be seen as an entropy-regularized variation of the scheme we proposed Eq.~\eqref{eq:ae_vi}. Now, taking the limit as the temperature goes to zero, we have
\begin{equation}
    \lim_{\tau\rightarrow 0} \Delta_b(s,a) = \beta b(s,a) + \max_{a'}(-\beta b(s,a')) = \beta (b(s,a) - \min_{a'} b(s,a')).
\end{equation}
Assuming moreover that the bonus is a prediction-based bonus, well-trained, meaning that $\min_{a'} b(s,a') = 0$ for any state $s$ in the dataset (as there is an action associated to it), we can rewrite the limiting case as Eq.~\eqref{eq:ae_vi}:
\begin{equation}
    \begin{cases}
        \pi_{k+1} = \argmax_\pi \langle \pi, Q_k - b \rangle 
        \\
        Q_{k+1} = r + \gamma P\langle \pi_{k+1}, Q_k - b \rangle
    \end{cases}.
\end{equation}
\looseness=-1
Thus, we can see the proposed anti-exploration VI scheme as a limiting case, when the temperature goes to zero, of a KL-regularized scheme that regularize the learned policy towards a behaviorial policy constructed from the bonus, assigning higher probabilities to lower bonuses. This derivation is reminiscent to advantage learning~\citep{bellemare2016increasing} being a limiting case of KL-regularized VI~\citep{vieillard2020leverage} in online RL~\citep{vieillard2020munchausen}.

\section{A practical approach}
\label{sec:algorithm}

\begin{figure}[h!]
    \includegraphics[width=\linewidth]{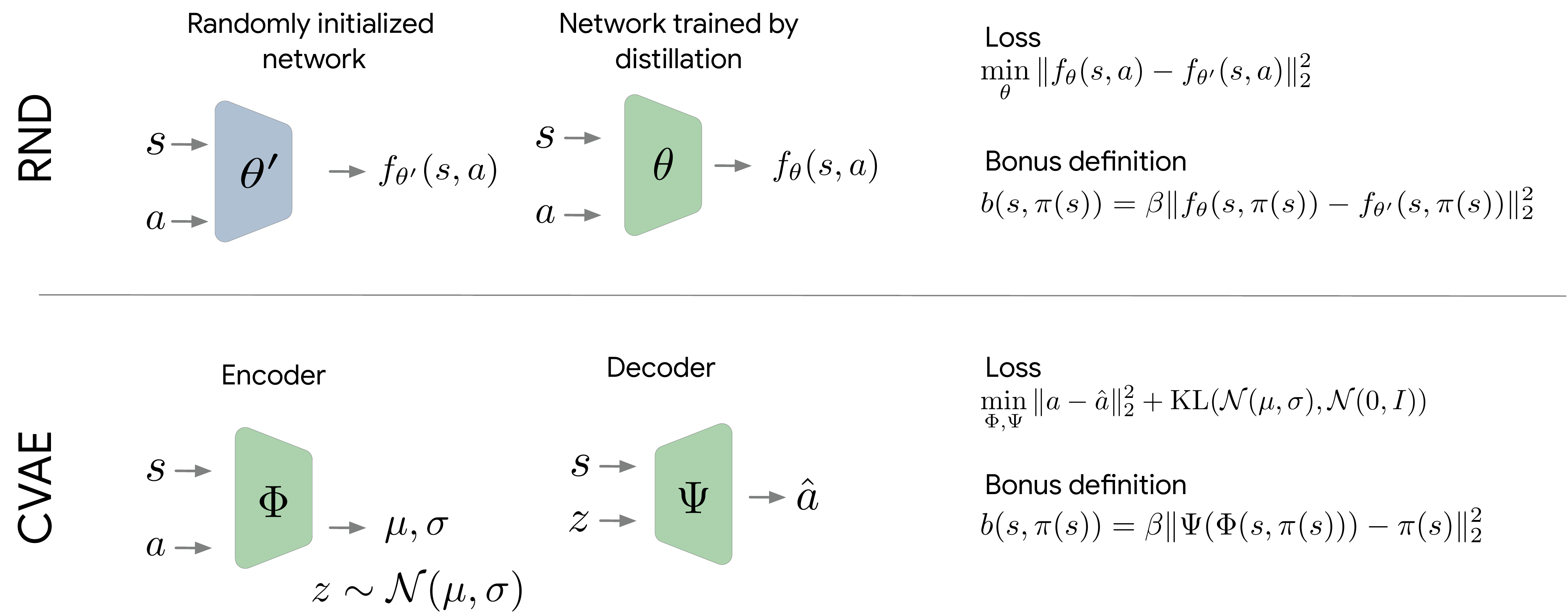}
    \caption{Illustration of RND and CVAE networks, losses and inferred anti-exploration bonuses.}
    \label{fig:networks}
\end{figure}

In principle, any (VI-based) RL agent could be combined with any exploration bonus for providing an anti-exploration agent for offline RL. For example, in a discrete action setting, one could combine DQN, described briefly in Sec.~\ref{sec:background}, with RND~\citep{burda2018exploration}. The RND exploration bonus is defined as the prediction error of encoded features as shown in Figure \ref{fig:networks}: a prediction network is trained to predict the output of a fixed random neural network denoted as target network, ideally leading to higher prediction errors when facing unfamiliar states. 

\looseness=-1
Here, we will focus on a continuous action setting. To do so, we consider the TD3 actor-critic as the RL agent, briefly described in Sec.~\ref{sec:background}. For the exploration bonus, RND would be a natural choice. However, in our experiments, RND performed poorly. This might be because RND was introduced for discrete action spaces and image-based observations, for which we know that random CNNs capture meaningful features. However, extending it to continuous control is not straightforward. Our experiments in Section \ref{sec:experiment_bonus} suggest that RND is not discriminative enough to separate state-actions in the dataset from others. Therefore, we introduce a bonus based on the reconstruction error of a variational autoencoder (that could be also useful in an exploration setting, \textit{in fine}).

\looseness=-1
\textbf{TD3.} We described briefly TD3 before, from a VI viewpoint. It indeed comes with additional particularities, that are important for good empirical performance. The noise (the standard deviation of the Gaussian policy) is not necessarily the same in the actor critic losses, a twin-critic is considered for reducing the overestimation bias when bootstrapping, and policy updates are delayed. We refer to \citet{fujimoto2018addressing} for more details. We use the classic TD3 update, except for the additional bonus term:
\begin{align}
    J_{\text{td3, actor, }\color{red}b}(\theta) &= \hat{\E}_{s\in\mathcal{D}}[\hat{\E}_{\epsilon\sim\mathcal{N}(0,\sigma I)}[Q_{\bar{\omega}}(s,\mu_\theta(s) + \epsilon) {\color{red}- b(s, \mu_\theta(s) + \epsilon)}]],
    \label{eq:loss_actor}
    \\
    J_{\text{td3, critic, }\color{red}b}(\omega) &= \hat{\E}_{\mathcal{D}}[\hat{E}_{a'\sim\pi_\theta(\cdot|s')}[(r + \gamma Q_{\bar{\omega}}(s',a') {\color{red} - b(s',a')} - Q_\omega(s,a))^2].
    \label{eq:loss_critic}
\end{align}

\textbf{CVAE.} 
The bonus we use for anti-exploration is based on a Conditional Variational Autoencoder (CVAE)~\citep{sohn2015learning}. The Variational Autoencoder (VAE) was first introduced by \citet{VAE_Kingma2014}. The model consists of two networks, the encoder $\Phi$ and  the decoder $\Psi$. The input data $x$ is encoded to a latent representation $z$ and then the samples $\hat{x}$ are generated by the decoder from the latent space.  
Let us consider a dataset, $X =\{x^1, . . . , x^N\},$ consisting of N i.i.d. samples. We assume that the data were generated from low dimensional latent variables $z$. VAE performs density estimation on $P(x, z)$ to maximize the likelihood of the observed training data $x$: 
$\log P(x)=\sum_{i=1}^{N}{\log P(x^{i})}$. 
Since this marginal likelihood is difficult to work with directly for non-trivial models, instead a parametric inference model $\Psi(z | x)$ is used to optimize the variational lower bound on the marginal log-likelihood: 
$\mathcal{L}_{\Phi, \Psi} = E_{P(z | x)}[\log \Phi(x | z)] - \kl{\Psi(z | x)}{\Phi(z)}$. 
%
A VAE optimizes the lower bound by reparameterizing $\Psi(z | x)$ \citep{VAE_Kingma2014, rezende14}. The first term of $\mathcal{L}$ corresponds to the reconstruction error and the second term regularizes the distribution parameterized by the encoder $\Phi(x | z)$ to minimize the KL divergence from a chosen prior distribution, usually an isotropic, centered Gaussian. 
In our problem formulation, given a dataset $\dataset$, we use a conditional variational autoencoder to reconstruct actions conditioned on the states. Hence, $\mathcal{L}_{\Phi, \Psi}$ can be rewritten as:
\begin{equation}
\mathcal{L}_{\Phi, \Psi} = E_{P(z | s, a)}[\log \Phi(a | s, z)] - \kl{\Psi(z | s, a)}{\Phi(z | s)}.
\label{eq:loss_cvae}
\end{equation}

\textbf{Summary.}  
Our specific instantiation of the idea of anti-exploration for offline RL works as follows. Given a dataset of interactions, we train a CVAE to predict actions conditioned on states (see Alg.~\ref{alg:cvae}). For any given state-action pair, we define the bonus as being the scaled prediction error of the CVAE,
\begin{equation}
    b(s,a) = \beta \norm{a - \Psi(\Phi(s, a))}_2^2, \label{eq:bonus}
\end{equation}
with $\beta$ the scale parameter. This bonus modifies the TD3 losses,  run over the fixed dataset (Alg.~\ref{alg:act-cri}).
 
\begin{algorithm}
    \begin{algorithmic}[1]
    \floatname{algorithm}{Procedure}
    \STATE{Initialize CVAE networks $\Phi$ and $\Psi$} 
    \FOR {step $i = 0$ to $N$}
    \STATE{Sample a minibatch of $k$ state-action pairs $\{(s_t, a_t), t=1,..,k \}$ from $\dataset$}
    \STATE{Train $\Phi$ and $\Psi$ using $\mathcal{L}_{\Phi, \Psi}$, see Eq.~\eqref{eq:loss_cvae}}
    \ENDFOR
    \end{algorithmic}
    \caption{CVAE training.}
    \label{alg:cvae}
\end{algorithm}

\begin{algorithm}
    \begin{algorithmic}[1]
    \floatname{algorithm}{Procedure}
    \STATE{Initialize policy $\pi_\theta$, action-value network $Q_{\omega}$ and target network $Q_{\bar{\omega}}$, $Q_\omega$ } 
    \FOR {step $i = 0$ to $N$}
    \STATE{Sample a minibatch of $k$ transitions $\{(s_t, a_t, r_t, s_{t+1}), t=1,..,k \}$ from $\dataset$}
    \STATE{For each transition, decode the action and computer the bonus, see Eq.~\eqref{eq:bonus}}
    \STATE{Update critic: gradient step on $J_{\text{td3, critic, }\color{red}b}(\omega)$, see Eq.~\eqref{eq:loss_critic}}
    \STATE{Update actor: gradient step on $J_{\text{td3, actor, }\color{red}b}(\theta)$, see Eq.~\eqref{eq:loss_actor}}
    \STATE{Update target network $Q_{\bar{\omega}} := Q_\omega$}
    \ENDFOR
    \end{algorithmic}
    \caption{Modified TD3 training.}
    \label{alg:act-cri}
\end{algorithm}

\section{Experiments}
\label{sec:experiments}
After describing the experimental setup and considered datasets, we first evaluate the discriminative power of the CVAE-based anti-exploration bonus in identifying OOD state-action pairs, comparing it to the one of the more natural (in an exploration context, at least) RND (Sec.~\ref{sec:experiment_bonus}). Then, we compare the proposed approach to prior offline RL methods on a range of hand manipulation and locomotion tasks with multiple data collection strategies \citep{d4rl}. 

\paragraph{Experimental setup.} 
\label{sec:experiment_setup}
We focus on locomotion and manipulation tasks from the D4RL dataset \citep{d4rl}. 
Along with different tasks, multiple data collection strategies are also considered for testing the agent's performance in complex environments. 

First, for the locomotion tasks, the goal is to maximize the traveled distance. For these tasks, the datasets are: \texttt{random}, \texttt{medium-replay}, \texttt{medium} and lastly \texttt{medium-expert}.
\texttt{Random} consists of transitions collected by a random policy. \texttt{Medium-replay} contains the first million transitions collected by a SAC agent \citep{haarnoja18b} trained from scratch on the environment. \texttt{Medium} has transitions collected by a policy with sub-optimal performance. Lastly, \texttt{medium-expert} is build up from transitions collected by a near optimal policy next to transitions collected by a sub-optimal policy. 

\looseness=-1
Second, the hand manipulation tasks require controlling a 24-DoF simulated hand in different tasks of hammering a nail, opening a door, spinning a pen, and relocating a ball \citep{rajeswaran2017learning}. These tasks are considerably more complicated than the gym locomotion tasks with higher dimensionality. The following datasets were collected on hand manipulation tasks: \texttt{human}, \texttt{cloned}, and \texttt{expert}. The \texttt{human} dataset is collected by a human operator. \texttt{Clone} contains transitions collected by a policy trained with behavioral cloning interacting with the environment next to initial demonstrations. Lastly, \texttt{expert} is build up from transitions collected by a fine-tuned RL policy interacting in the environment.  

\paragraph{Anti-exploration bonus.} 
\label{sec:experiment_bonus}
We analyze the quality of the learned bonus in detail for different algorithms. In particular, we are interested in the capability of the anti-exploration bonus to separate state-action pairs in the dataset from others. However, even though it is straightforward to have positive examples (these are those in the dataset), it is much more difficult to define what negative examples are (otherwise, the bonus could be simply trained using a binary classifier). In these experiments, for a state-action pair $(s, a) \in \dataset$, we define an  OOD action $\Tilde{a}$ in three different ways.
First, we consider actions uniformly drawn from the action space, $\Tilde{a} \sim \U(\actions)$. Second, we consider actions from the dataset perturbed with Gaussian noise, $\Tilde{a} = a + \gamma \mathcal{N}(0, I)$. Third, we consider randomly shuffled actions (for a set of state-action pairs, we shuffle the actions and not the states, which forms new pairs considered as negative examples).

We investigate the discriminative power of the bonus in distinguishing OOD state-action pairs for two different models, namely RND (as it would be a natural choice in an exploration context, at least in a discrete action setting) and CVAE (the proposed approach for learning the bonus). 

As for the case of RND \citep{burda2018exploration}, state-action pairs are passed to the target network $f$ and prediction network $f'$. The prediction network is trained to predict the encoded feature from the target network given the same state-action pair, i.e., minimize the expected MSE between the encoded features. All the implementation details are provided in the Appendix. The bonus is defined as the prediction error of encoded features: $b(s, a) := \beta \norm{f(s, a) - f'(s, a) }_2^2$.
In the CVAE model, a state-action pair $(s, a)$ is concatenated and encoded to a latent representation $z$. This latent representation $z$ next to state $s$ is passed to the decoder to reconstruct action $a$. Both encoder and decoder consist of two hidden layers of size 750 with the latent size set to 12. We provide further details on the implementation in the Appendix. The bonus is defined as the reconstruction error of the actions: $b(s, a) := \beta \norm{a - \Psi(\Phi(s, a))}_2^2$.

\begin{figure}[h!]
    \includegraphics[width=\linewidth]{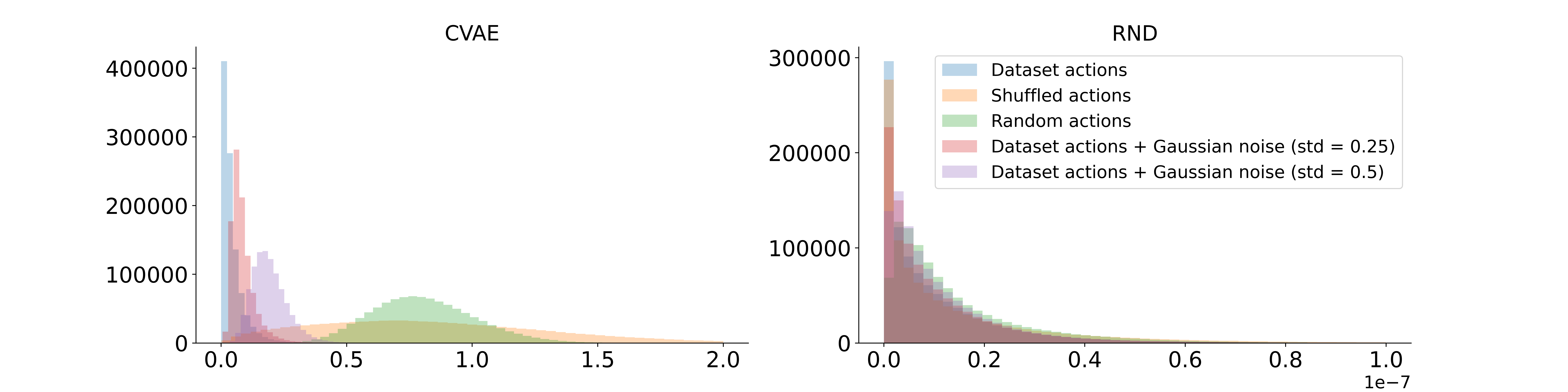}
    \caption{Visualization of the histogram of the reconstruction error for the entire dataset of walker2d-medium for RND and CVAE. The reconstruction error is computed for the original dataset state-action pairs (blue) as well as different perturbations of the actions: randomly permutated actions over the dataset (orange), random actions (green), original actions to which is added Gaussian noise with different standard deviations (red, purple). Results are similar for other datasets.}
    \label{fig:bonus_1}
\end{figure}

Histograms of the bonus for OOD state-action pairs are compared with those in the dataset and visualized in Figure \ref{fig:bonus_1}. For these experiments, the state was fixed, and the bonus was derived for different kind of OOD actions. The results are shown for CVAE and RND models in the left and right figures, respectively. As we expected, for the CVAE model, the bonus is mostly higher for shuffled, random, and noisy actions in contrast to the actions in the dataset. Moreover, the bonus gets higher as we increase the variance of the added noise. However, there is not much difference between the bonus for actions in the dataset and others in the RND model and it performs poorly in identifying OOD actions. The CVAE model, on the other hand, is discriminative enough to separate state-action pairs in the dataset from others. In fact, in the novelty detection task where the goal is to identify outliers, autoencoders were shown to be beneficial \citep{abati2019latent, pidhorskyi2018generative, xu2015learning, zhou2017anomaly}.

\paragraph{Performance on D4RL datasets.} 
\label{sec:exp_d4rl}
Now that we have assessed the efficiency of CVAE in discriminating OOD state-action pairs from the ones in the dataset, we combine it with TD3 and assess the performance of the resulting offline RL agent on the D4RL datasets depicted above. We compare to the model-free approaches BEAR~\citep{kumar2019stabilizing}, BRAC~\citep{BRAC}, AWR~\citep{peng2019advantage}, BCQ~\citep{fujimoto2019off} and CQL~\cite{cql}, the later providing state-of-the-art results on these tasks.

The architecture of the TD3 actor and critic consists of a network with two hidden layers of size 256, the first layer has a tanh activation and the second layer has an elu activation. The actor outputs actions with a tanh activation, which is scaled by the action boundaries of each environment.
Except from the activation functions, we use the default parameters of TD3 from the authors implementation, and run $10^6$ gradient steps using the Adam optimizer, with batches of size 256. All implementation details are provided in the Appendix.

In online RL, it is quite standard to scale or clip rewards, which can be hard as the reward range may not be known a priori. In an offline setting, all rewards that will be used are in the dataset, so their range is known and it is straightforward to normalize them. Therefore, we scale the reward of each environment such that it is in $[0,1]$. This is important for having a scale parameter $\beta$ for the bonus not too much dependent from the task. Indeed, we have shown that our anti-exploration idea ideally optimizes for $\E_\pi[\sum_{t\geq 0} \gamma^t(r(s_t,a_t)-b(s_t,a_t))]$. Thus, the scale of the bonus should be consistent with the scale of the reward, which is easier with normalized rewards.

TD3 allows for different levels of noises in the actor and critic losses, which deviates a bit from the VI viewpoint. We adopt a similar approach regarding the scale of the bonus, as it provides slightly better results. We allow for different scales $\beta_a$ and $\beta_c$ for the bonus in respectively the actor loss and critic loss. 
We execute a hyperparameter search over the bonus weights over $\beta_a, \beta_c \in \{ 0.1, 0.5, 1, 5, 10\}$. For all locomotion tasks, we select the pair of scale factors providing the best result (so a single pair for all tasks, not one per task), and the same for manipulation tasks. For locomotion $\beta_a = 5$ and $\beta_c = 1$, and for manipulation tasks $\beta_a = \beta_c = 10$ were chosen. 

We show the performance of the proposed approach on the D4RL datasets in Table \ref{tab:performance} (depicted there as TD3-CVAE). We report the average and standard deviation of the results over 10 seeds, each being evaluated on 10 episodes. On average, it is competitive with CQL and outperforms others in locomotion tasks. On the hand manipulation tasks, CVAE outperforms all other methods. 
Notice that all considered baselines are model-free. Better results can be achieved by model-based approaches \citep{yu2021combo, yu2020mopo}, at least on the locomotion tasks. Notice that the model-free or model-based aspect is orthogonal to our core contribution, the idea of anti-exploration could be easily combined to a model-based approach. We let further investigations of this aspect for future works.

\begin{table*}[h!]
  \centering
\begin{tabularx}{\linewidth}{byyxxyyya}
\toprule
Algorithm &                        BC &  BEAR &  BRACp &BRACv &   AWR &    BCQ &    CQL &  TD3-CVAE \\
\midrule
halfcheetah-random        &    2.1 &  25.1 &    24.1 &  31.2 &   2.5 &    2.2 &   \textbf{35.4} &   28.6 $\pm$   2.0 \\
walker2d-random           &    1.6 &   \textbf{7.3} &    -0.2 &   1.9 &   1.5 &    4.9 &    7.0 &    5.5 $\pm$   8.0 \\
hopper-random             &    9.8 &  11.4 &    11.0 &  \textbf{12.2} &  10.2 &   10.6 &   10.8 &   11.7 $\pm$   0.2 \\
halfcheetah-medium        &   36.1 &  41.7 &    43.8 &  \textbf{46.3} &  37.4 &   40.7 &   44.4 &   43.2 $\pm$   0.4 \\
walker2d-medium           &    6.6 &  59.1 &    77.5 &  \textbf{81.1} &  17.4 &   53.1 &   79.2 &   \underline{68.2 $\pm$   18.7} \\
hopper-medium             &   29.0 &  52.1 &    32.7 &  31.1 &  35.9 &   54.5 &  \textbf{58.0} &    \underline{55.9 $\pm$   11.4} \\
halfcheetah-med-rep &   38.4 &  38.6 &    45.4 &  \textbf{47.7} &  40.3 &   38.2 &   46.2 &   45.3 $\pm$   0.4 \\
walker2d-med-rep    &   11.3 &  19.2 &    -0.3 &   0.9 &  15.5 &   15.0 &   \textbf{26.7} &   15.4 $\pm$   7.8 \\
hopper-med-rep      &   11.8 &  33.7 &     0.6 &   0.6 &  28.4 &   33.1 &   \textbf{48.6} &   \underline{46.7 $\pm$   17.9} \\
halfcheetah-med-exp &   35.8 &  53.4 &    44.2 &  41.9 &  52.7 &   64.7 &   62.4 &   \textbf{86.1 $\pm$    9.7} \\
walker2d-med-exp    &    6.4 &  40.1 &    76.9 &  81.6 &  53.8 &   57.5 &  \textbf{111.0} &   84.9 $\pm$   20.9 \\
hopper-med-exp      &  \textbf{111.9} &  96.3 &     1.9 &   0.8 &  27.1 &  110.9 &   98.7 &  \underline{111.6 $\pm$   2.3} \\
\midrule
Mean performance          &   25.0 &  39.8 &    29.8 &  31.4 &  26.8 &   40.4 &   \textbf{52.3} &   \underline{50.3 $\pm$   8.3} \\
\midrule
pen-human       &   34.4 &   -1.0 &  8.1 &  0.6 &   12.3 &   \textbf{68.9} &   37.5 &   \underline{59.2 $\pm$  14.3} \\
hammer-human    &    1.5 &    0.3 &  0.3 &  0.2 &    1.2 &    0.5 &    \textbf{4.4} &    0.2 $\pm$   0.0 \\
door-human      &    0.5 &   -0.3 & -0.3 & -0.3 &    0.4 &   -0.0 &    \textbf{9.9} &    0.0 $\pm$   0.0 \\
relocate-human  &    0.0 &   -0.3 & -0.3 & -0.3 &   -0.0 &   -0.1 &    \textbf{0.2} &   -0.0 $\pm$   0.0 \\
pen-cloned      &   \textbf{56.9} &   26.5 &  1.6 & -2.5 &   28.0 &   44.0 &   39.2 &   \underline{45.4 $\pm$  25.5} \\
hammer-cloned   &    0.8 &    0.3 &  0.3 &  0.3 &    0.4 &    0.4 &    \textbf{2.1} &    0.3 $\pm$   0.1 \\
door-cloned     &   -0.1 &   -0.1 & -0.1 & -0.1 &    0.0 &    0.0 &    \textbf{0.4} &    0.0 $\pm$   0.1 \\
relocate-cloned &   \textbf{-0.1} &   -0.3 & -0.3 & -0.3 &   -0.2 &   -0.3 &   -0.1 &   -0.2 $\pm$   0.0 \\
pen-expert      &   85.1 &  105.9 & -3.5 & -3.0 &  111.0 &  \textbf{114.9} &  107.0 &  \underline{112.3 $\pm$  21.9} \\
hammer-expert   &  125.6 &  127.3 &  0.3 &  0.3 &   39.0 &  107.2 &   86.7 &  \textbf{128.9 $\pm$   1.5} \\
door-expert     &   34.9 &  \textbf{103.4} & -0.3 & -0.3 &  102.9 &   99.0 &  101.5 &   59.4 $\pm$  34.7 \\
relocate-expert &  101.3 &   98.6 & -0.3 & -0.4 &   91.5 &   41.6 &   95.0 &  \textbf{106.4 $\pm$   5.0} \\
\midrule
Mean performance   &   36.7 &   38.3 &  0.4 & -0.4 &   32.2 &   39.6 &  \textbf{40.3} &   \textbf{42.6 $\pm$   8.6} \\
\bottomrule
\end{tabularx}
\caption{\label{tab:performance} Evaluation of CVAE. We report the results of the baselines using performance results reported by \citet{d4rl}, which do not incorporate standard deviation of performances, since the numbers are based on 3 seeds. In our case we use 10 seeds, following recommendations from \citet{henderson}, and evaluate on 10 episodes for each seed before reporting average and standard deviations of performance. We bold best average performance and underline the performance when whithin one standard deviation of the best performance.}
\end{table*}

\section{Related work}
\looseness=-1
Here, we briefly discuss prior work in offline RL and how our proposed approach differs. As discussed in Section \ref{sec:background}, offline RL suffers from extrapolation error cause by distribution mismatch. Recently, there has been some progress in offline RL to tackle this issue. They can be roughly categorized into policy regularization-based and uncertainty-based methods. The former constrains the learned policy to be as close as possible to the behavioral policy in the dataset. The constraint can be implicit or explicit and ensures that the value function approximator will not encounter out-of-distribution (OOD) state-actions. The difference lies in how the measure of closeness is defined and enforced. Some of the most common measures of closeness are the KL-divergence, maximum mean discrepancy (MMD) distance, or Wasserstein distance \citep{BRAC}. In AWR \citep{peng2019advantage}, CRR \citep{wang2020critic} or AWAC \citep{nair2020accelerating}, the constraint is introduced implicitly by incorporating a policy update that keeps it close to the behavioral policy. Furthermore, the constraints can be enforced directly on the actor update or the value function update. In BRAC, \citet{BRAC}  investigated different divergences and choices of value penalty or policy regularization.  In BEAR \citet{kumar2019stabilizing}  argue that restricting the support of the learned policy to the support of the behavior distribution is sufficient, allowing more flexibility and a wider range of actions to the algorithm. 
An alternative approach to alleviate the effects of OOD state-actions is to make the value function approximators robust to those state-actions. The aim is to make the target value function for OOD state-actions more conservative \citep{cql, buckman2020importance, laroche2019safe}.

\looseness=-1
We proposed an alternative approach inspired from exploration. Contrary to the online setting, an offline agent should avoid drifting from the distribution of the dataset. Hence, we formulate this as an ``anti-exploration'' problem where OOD state-action pairs are penalized. Focusing on bonus-based approaches~\citep{burda2018exploration, pathak2017curiosity, burda2018large, BellemareSOSSM16}, we define an anti-exploration bonus to be subtracted from the reward. This general idea opens up new research directions for adapting any exploration strategy to the offline setting. Moreover, with minimal assumptions, our proposed method can be linked to the regularization-based methods, somehow drawing a bridge between regularization- and uncertainty-based methods. 

\section{Conclusion}
We proposed an intuitive and straightforward approach to offline RL. We constraint the policy to take state-action pairs within the dataset to avoid extrapolation errors. To do so, the core idea is to \emph{subtract} a prediction-based \emph{exploration bonus} from the reward (instead of adding it for exploration). We theoretically showed the connection of the proposed method with the regularization-based approaches.
Instantiating this idea with a CVAE-based bonus and the TD3 agent (which is a possibility among many others), we reached state-of-the-art performance on D4RL datasets. Our approach is quite versatile and general. As such, an interesting research direction would be to combine it with orthogonal improvements to offline RL, such as considering a model-based agent~\citep{yu2020mopo}.


\bibliographystyle{abbrvnat}
\bibliography{main}
\newpage

\appendix
\section{Implementation details}
In this section, we provide a detailed description of the experimental study. 

\paragraph{Preprocessing.} We use D4RL datasets \citep{d4rl}. %

We normalize the rewards such that they are in $[0,1]$,
\begin{equation}
    r \leftarrow \frac{r - \min_{r \sim \dataset} r}{\max_{r \sim \dataset} r - \min_{r \sim \dataset} r}.
\end{equation}
This makes the reward range consistent across different environments, without changing the set of optimal policies. 

\paragraph{CVAE.} Both encoder  $\Phi$ and decoder  $\Psi$ consist of two hidden layers of size 750 with relu activation. The encoder takes the concatenation of state and action as input and encodes it to the mean $\mu$ and standard deviation $\sigma$ of a Gaussian distribution $\mathcal{N}(\mu, \sigma)$. The latent representation $z$ is sampled from the Gaussian, is further concatenated with the state, and passed to the decoder, which outputs an action. We train the CVAE for 50'000 gradient steps with batch size 100 and hyperparameters defined in Table \ref{tab:HP_cvae} with respect to the marginal log likelihood loss:
\begin{equation}
\mathcal{L}_{\Phi, \Psi} = E_{P(z | x)}[\log \Phi(x | z)] - \eta \kl{\Psi(z | x)}{\Phi(z)}. 
\end{equation}

\begin{table*}[h!]
  \centering
\caption{\label{tab:HP_cvae} Default Hyperparameters for CVAE }
\begin{tabularx}{0.5\linewidth}{cc}
\toprule
Hyperparameter & Value \\
\midrule
Optimizer & Adam \\
Learning Rate & $10^{-4}$ \\
Batch Size & 100 \\ 
Latent Size & 12 \\
$\eta$ & 0.5\\
Normalized Observations & False\\
\bottomrule
\end{tabularx}
\end{table*}

\paragraph{Agent.}  We re-implemented the TD3 agent from \citet{fujimoto2019off} with the default hyperparameters.
The architecture of the TD3 actor and critic consists of a network with two hidden layers of size 256, the first layer has a tanh activation and the second layer has an elu activation. The actor outputs actions with a tanh activation, which is scaled by the action boundaries of each environment.
We used the Adam optimizer for both the actor and the critic with a learning rate of $3.10^{-4}$, and trained them using batch of transitions of size 256 sampled uniformly in $\dataset$, for $500'000$ gradient steps. 

We execute a hyperparameter search for both the locomotion environments and the hand manipulation environments over the bonus weights $\beta_a, \beta_c \in \{ 0.1, 0.5, 1, 5, 10\}$. The best combination for the average normalized performance on the tasks (averaged over 3 seeds) is selected. We re-ran the best combination of hyperparameters for 10 seeds and report results averaged over the 10 seeds and 10 evaluation episodes per seed. For locomotion tasks the best set of weights was $\beta_a = 5$ and $\beta_c = 1$, and for manipulation tasks $\beta_a = \beta_c = 10$. 

\section{Additional empirical analysis of anti-exploration bonus}

\begin{figure}[h!]
    \centering
    \begin{subfigure}
    \centering
    \includegraphics[width=\linewidth]{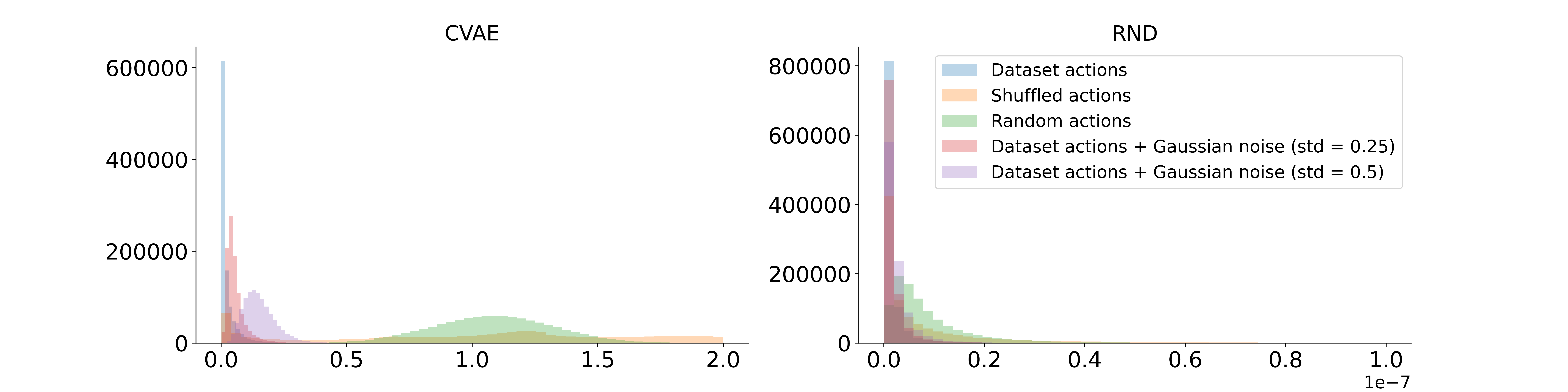}
    \end{subfigure}
    \begin{subfigure}
    \centering
    \includegraphics[width=\linewidth]{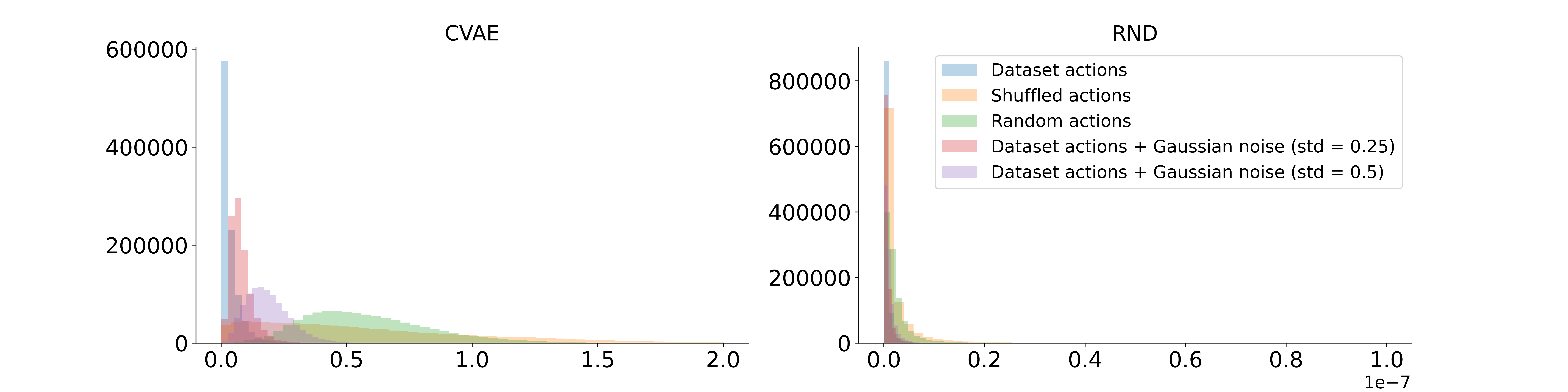}
    \end{subfigure}
    \caption{Visualization of the histogram of the reconstruction error for the entire dataset of halfcheetah-medium and hopper-medium in the top and bottom figures respectively. The reconstruction error is computed for the original dataset state-action pairs (blue) as well as different perturbations of the actions: randomly permutated actions over the dataset (orange), random actions (green), original actions to which is added Gaussian noise with different standard deviations (red, purple).}
    \label{fig:bonus_2}
\end{figure}

We empirically analyze the quality of the learned bonus in detail for other environments. In particular, we are interested in the discriminative power of the bonus in separating state-action pairs in the dataset from others. For these experiments, the state was fixed, and the bonus was derived for different types of OOD actions. This is the same experiment as for Fig.~\ref{fig:bonus_1}, but on more datasets. The histogram of the reconstruction error for 
``halfcheetah-medium'' and ``hopper-medium'' for both RND and CVAE models are shown in Fig.~\ref{fig:bonus_2}.

\begin{figure}[h!]
    \centering
    \includegraphics[width=\linewidth]{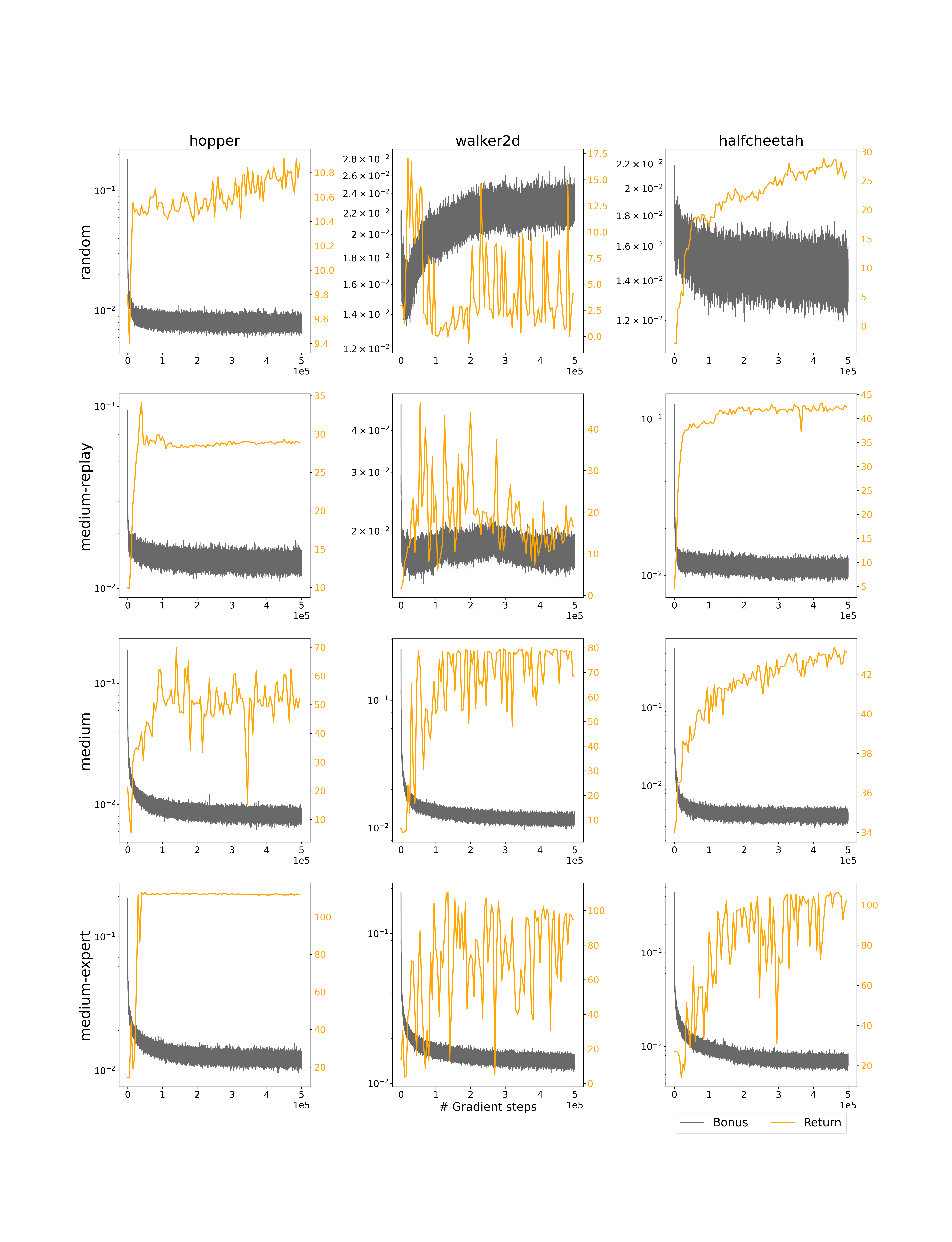}
    \caption{Visualization of average bonus (grey) and return (yellow) during training for different environments.}
    \label{fig:bonus_3}
\end{figure}

Additionally, we show on Figure~\ref{fig:bonus_3} the average anti-exploration bonus for both actor and critic during training the agent policy, as well as the return of the learnt policy. Ideally, the agent policy should get trained to select state-action pairs with lower anti-exploration bonuses. As shown in the figure, the average bonus smoothly decreases, showing the agent is avoiding taking OOD state-action pairs. Notice that the agents do not only learn to take low-bonus actions, but also learn to maximize cumulative rewards. Otherwise, it would not perform better than behavioral cloning.

\end{document}